\documentclass[10pt]{article}
\usepackage[T1]{fontenc}
\usepackage{lmodern}
\input{glyphtounicode}
\usepackage[margin=1in]{geometry}
\usepackage{amsmath,amssymb,booktabs,hyperref,xcolor,enumitem}
\usepackage[numbers]{natbib}
\usepackage{pgfplots}
\pgfplotsset{compat=1.17}

\title{Baseline Shape Decides the Verdict: A Controlled Re-Examination\\of Ternary Language Models at 60K Parameters}
\author{Gautam Veldanda\\\small Independent Researcher \quad \texttt{veldandax@gmail.com}}
\date{}

\begin{document}
\maketitle

\begin{abstract}
Ternary (1.58-bit) weights are attractive for microcontroller-class language models, but the
sub-1M-parameter regime is supported mainly by isolated, single-seed comparisons. A prominent
example reports that a routed ternary block (local convolution, diagonal state-space model and
sparse attention mixed by a per-token router) beats a parameter-matched full-precision transformer
by 22\% in perplexity at 60K parameters, and attributes this to architectural inductive bias. We
re-run that comparison under one fixed recipe with three seeds per cell, 98 runs in total, on a
single laptop. The models are byte-level, so budgets below are byte counts, and every relative gap
is divided by the reference system named beside it. Four results follow. (i) Baseline \emph{shape}
dominates the comparison: at a 16M-byte budget, param-matched transformers span 2.835--3.477
nats/byte purely by depth/width choice (a 22.6\% span, far larger than any architecture effect we
measure at that budget), and at this controlled budget the routed model's advantage disappears
against the best-shaped transformer, so the published margin is at least partly a baseline-shape
effect; the ordering of shapes reverses with budget, so a single fixed shape cannot be trusted. (ii) With a 130M-byte budget the routed model does win, by
22.2--24.0\% over each of the three transformer shapes we evaluate there, but a plain gated
diagonal-SSM block beats it by a further 9.1\% in full precision, and the routed model's own router
puts most of its weight on its recurrent pathway, so the gain does not require routing. (iii) The
ternary penalty differs sharply by architecture at the larger budget (+5.3\% for the best
transformer shape against +19.5\% routed and +28.1\% gated SSM), but we cannot attribute that
difference to architecture alone: our transformers keep learned positional embeddings in full
precision, 11--22\% of their parameters, so they are less quantized than the models they are
compared with. (iv) A 90/10 full-precision-then-ternary schedule beats all-ternary training here,
but only with a stage-2 learning rate about 10$\times$ the pretraining peak; at a conventional
fine-tuning rate the same schedule looks 15.3\% worse, reversing the conclusion. The from-scratch
baseline was not itself learning-rate tuned, which bounds (iii) and (iv). All code, corpus manifests
and run logs are released.\footnote{\url{https://github.com/veldanda/ByteLM}, release tag
\texttt{p1-v1}. The tag contains the training code, the parameter-matched configs, the corpus
manifest with SHA-256, and one directory of logs and checkpointed metrics per run.}
\end{abstract}

\section{Introduction}
Running a generative language model on a microcontroller (MCU) leaves a budget of tens to
hundreds of kilobytes of flash and SRAM. Ternary weights, $w\in\{-\alpha,0,+\alpha\}$ at
$\log_2 3\approx1.58$ bits, are a natural fit: they shrink storage by up to
$\sim$20$\times$ relative to FP32 and replace multiplications with additions
\citep{ma2024era}. Whether ternary training is \emph{competitive} at these sizes, however,
is poorly understood. The most careful low-bit scaling studies (ParetoQ
\citep{liu2025paretoq} and Spectra \citep{kaushal2024spectra,vaidhya2025spectra11}) begin
at 125M and 99M parameters respectively, and the recent sub-100M study of
\citet{thomassen2026schedule} considers integer bit-widths (8/6/4) on transformers only.

Below 10M parameters, the evidence is largely anecdotal. Atome-LM \citep{atomelm2026} reports
that a routed ternary hybrid beats a parameter-matched FP32 transformer by 22\% perplexity at
60K parameters, but loses by $\sim$11\% at 944K, and interprets this as the architecture's
inductive bias ``substituting for capacity at small scale and constraining it at larger
scale''. If true, this would be an important design rule for edge models: use strong
structural priors only in the smallest tier. But the claim rests on two single-seed
comparisons, and, as we show in \S\ref{sec:audit}, the 60K and 944K comparisons were run
under very different training regimes.

Re-running that comparison under one recipe with three seeds gives a different picture, and the
first reason is mundane. At a fixed 60K parameter budget, a transformer's depth/width ratio moves
validation loss by 22.6\%, further than replacing the architecture does, so the published result
depends on which single transformer shape is used as the baseline. Once the strongest shape is
used at each budget, two things remain: at a larger byte budget the routed block genuinely wins,
but a plain gated recurrence wins by more; and the measured cost of ternary weights differs by
architecture, shrinking with training for transformers and not for recurrent models, though
that difference is confounded by unequal quantization and by an untuned from-scratch ternary
baseline, as we set out in \S\ref{sec:limits}.
We therefore ask: \emph{at microcontroller scale, what actually accounts for reported architectural
advantages, and how does the ternary penalty behave as the training budget grows?}

\paragraph{Contributions.}
\begin{enumerate}[leftmargin=*,itemsep=2pt]
  \item \textbf{An audit of the published evidence} (\S\ref{sec:audit}): the 60K and 944K
  comparisons differ by ${\sim}650\times$ in training tokens and in corpus, schedule, optimizer
  settings and baseline FFN width, and the 944K comparison changes precision and architecture at once.
  \item \textbf{A controlled study at 60K}, 98 runs, three seeds per cell, one fixed recipe,
  token-matched full-precision anchors and a $2\sigma$ seed-noise criterion fixed before the results
  were inspected
  (\S\ref{sec:design}): baseline shape sensitivity, a three-family architecture comparison, the
  budget dependence of the ternary penalty, and a tuned FP-init recipe (\S\ref{sec:results}).
  \item \textbf{Two methodological warnings} that each reverse a headline: a fixed baseline shape
  (which flips the architecture verdict at 16M bytes), and an untuned stage-2 learning rate
  (which flips the FP-init verdict).
  \item \textbf{An open, laptop-scale release}: an identity quantizer that enables the first
  full-precision run of the routed architecture, plus exact re-implementations of two of its layers
  that make the study affordable (Appendix~\ref{app:kernels}); 98 runs in ${\sim}40$ device-hours
  on one MacBook Air.
\end{enumerate}

\section{Related Work}
\paragraph{Low-bit training and scaling.} BitNet b1.58 \citep{ma2024era} popularized ternary
weights with per-tensor absmean scaling and the straight-through estimator
\citep{bengio2013ste}. Spectra \citep{kaushal2024spectra} trains ternary transformers from
scratch from 99M to 3.9B parameters; Spectra~1.1 \citep{vaidhya2025spectra11} finds that
ternary models ``benefit more from increasing training data than from scaling model
parameters''. ParetoQ \citep{liu2025paretoq} compares 1--4-bit quantization-aware training
(QAT) from 125M to 8B parameters and finds that spending $\sim$90\% of the budget in full
precision before switching to QAT outperforms low-bit training from scratch, with $\leq$2-bit
models requiring more tokens to saturate. None of these studies go below $\sim$100M
parameters or vary architecture.

\paragraph{Sub-100M quantization.} \citet{thomassen2026schedule} runs a 720-run factorial of
INT8/INT6/INT4 QAT on 3M--350M transformers and finds the optimal learning-rate warmdown to be
bit-width-agnostic, with an INT4 regime boundary at 30--50M parameters and an INT6 penalty
that \emph{grows} with training length. Their study uses transformers only, integer
bit-widths and QAT from initialization; we study ternary weights, three architecture
families, and FP-initialized vs.\ from-scratch training, down to 60K parameters.
TernaryLM \citep{nargund2026ternarylm} trains a 132M ternary transformer on TinyStories with
full-precision embeddings and head.

\paragraph{Non-transformer low-bit models.} Ternary Mamba \citep{ganesaraja2026ternarymamba}
applies grouped QAT to a pretrained 1.3B Mamba-2 \citep{gu2023mamba}. Hybrid attention/SSM
designs such as Hymba \citep{dong2024hymba} target the $\geq$100M regime in full precision.
Atome-LM \citep{atomelm2026} is the only publicly released routed ternary hybrid at MCU scale we
are aware of.

\paragraph{Depth, width and the shape of a small model.} At a fixed parameter budget, depth and
width trade off against each other, and the optimum moves with the compute or data budget: this is
standard for large models \citep{hoffmann2022chinchilla,tay2022scaling} but is rarely controlled in
small-model comparisons, where a single ``param-matched baseline'' is usually reported. Our shape
sweep (\S\ref{sec:results}) shows the effect is large enough at 60K parameters to determine the
verdict of an architecture comparison, and that the best shape at one budget is the worst at
another.

\paragraph{Inductive bias and scale.} \citet{tay2022scaling} show across ten architectures
that ``the best performing model can fluctuate at different scales''. They vary model size in
full precision; we vary training budget and precision as well, and ask whether the
fluctuation reported at MCU scale is a size effect at all.

\section{Background: the Atome MCU block}\label{sec:bg}
Each block applies LayerNorm, then three pathways in parallel (a depthwise causal
convolution ($k{=}5$), a diagonal SSM $h_t = a\odot h_{t-1} + b\odot x_t$, $y_t = c\odot h_t$
with $a=\tanh(a_{\text{raw}})$, and single-head top-$k$ ($k{=}4$) causal attention), mixed by
a per-token softmax router, followed by a residual connection. There is no feed-forward
layer; channel mixing happens only in the attention value projection and the router.
Tokens are raw bytes (vocabulary 256) and there is no positional embedding.
All linear, convolution, embedding, unembedding and router weights are ternarized with a single
per-tensor absmean scale, $\tilde W=\alpha\,\mathrm{clip}(\mathrm{round}(W/\alpha),-1,1)$,
$\alpha=\mathrm{mean}|W|$, recomputed every forward pass with STE; the SSM's per-channel vectors
stay in FP32. Embedding and head are 13.9\% of parameters at 944K.

\section{Auditing the published comparison}\label{sec:audit}
Table~\ref{tab:audit} summarizes the conditions behind the two published data points,
reconstructed from the released run artifacts (\texttt{ab\_results.json},
\texttt{*.train.json}) and scripts at commit \texttt{be12186}.

\begin{table}[h]\centering\footnotesize
\caption{Conditions of the two published Atome-LM comparisons.}\label{tab:audit}
\begin{tabular}{@{}p{4.3cm}p{5.1cm}p{5.6cm}@{}}\toprule
 & 60K comparison & 944K comparison \\\midrule
Result (routed ternary vs.\ FP32 transformer) & 6.31 vs.\ 8.12 ppl (win) & 1.0545 vs.\ 0.9337 nats/byte (loss) \\[2pt]
Training tokens & $\approx$3M (3{,}000 steps $\times$ 16 $\times$ 64) & $\approx$2B (30{,}000 $\times$ 256 $\times$ 256) \\
Tokens per parameter & $\approx$50 & $\approx$2{,}100 \\
Corpus & 0.5\,MB TinyStories slice & $\sim$1.7\,GB TinyStories (reported) \\
Sequence length & 64 & 256 \\
LR schedule & constant $3\!\times\!10^{-4}$ & warmup 1000, cosine $3\!\times\!10^{-4}\!\to\!3\!\times\!10^{-5}$ \\
AdamW $(\beta_2$, weight decay$)$ & $(0.999,\ 0.01)$ (defaults) & $(0.95,\ 0.1)$ \\
Transformer FFN width & $1\times d$ & $4\times d$ \\
FP32 routed model trained? & no & no \\\bottomrule
\end{tabular}
\end{table}

Three observations follow.
(i) Scale is confounded with training budget and recipe: a strong structural prior that helps
in a 3M-token, constant-LR regime and is overtaken when a generic model receives 650$\times$ more
data would produce the same ``reversal'' with no size effect at all (\emph{H-budget}).
(ii) Neither comparison isolates precision from architecture, because no full-precision routed
model was trained; the released quantizer registry has no identity option.
(iii) The 60K transformer baseline uses FFN width $1\times d$, while the 944K baseline uses
$4\times d$. A fourth difference is worth naming: at 3M tokens over a 0.5\,MB corpus the 60K runs
make roughly six passes over their data, a regime in which memorization contributes to validation
loss, whereas our budgets stay under 0.07 epochs. Their reported 6.31 perplexity ($\approx1.84$
nats/byte) is far below anything we reach at 16M bytes (2.7--3.5), which is consistent with that
reading. We do not re-run their setup, so we cannot separate memorization from architecture there;
it is a further reason the two published points are not comparable to each other. We also note that the documented corpus-construction command produces at most
1\,MB of the TinyStories V2 validation file, whereas the results document states the full
train+valid set was used; we therefore build our own corpus with a checksum manifest and
re-run every comparison rather than relying on reported numbers.
We stress that this reconstruction was only possible because the authors released their run
artifacts, configurations and training scripts, which is more than most work at this scale
provides; the conditions above are stated in their own repository. Our point is not that their measurements are wrong (we do not re-run their exact 60K setup, and
have no reason to doubt the numbers), but that the two data points cannot be compared to each
other, and that the 60K comparison rests on one baseline shape (\S\ref{sec:results}).

\section{Experimental design}\label{sec:design}
\paragraph{Hypotheses.} (Outcomes in \S\ref{sec:results}; a further hypothesis on quantization
granularity, namely full-precision embedding/head and per-row scales, was specified but not
run, and is left to future work.)
\begin{description}[leftmargin=1em,itemsep=2pt]
  \item[H-budget.] The routed architecture's advantage over a transformer at fixed size shrinks
  or reverses as the training budget grows, independent of precision.
  \item[H-recipe.] Ternary models initialized from a full-precision checkpoint of the same
  architecture (ParetoQ-style) close part of the ternary--FP gap at sub-10M scale.
  \item[H-interaction.] The ternary penalty differs between architectures (precision
  $\times$ architecture interaction), beyond seed noise.
\end{description}

\paragraph{What was run (98 runs, 3 seeds per cell unless noted).}
All at 60K parameters, byte-level, on TinyStories V2.
\begin{itemize}[leftmargin=*,itemsep=1pt]
  \item \textbf{Main grid (24):} \{routed, transformer $3\times32$\} $\times$ \{FP32, ternary\}
        $\times$ \{16M, 130M bytes\}.
  \item \textbf{Shape sweep (36):} five param-matched transformer shapes ($1\times52$, $2\times36$,
        $3\times32$, $4\times28$, $6\times24$; all within 3\% of 60{,}800 parameters) at 16M, and
        the three strongest at 130M, in both precisions.
  \item \textbf{Third architecture (12):} gated diagonal-SSM block, both budgets, both precisions.
  \item \textbf{FP-init (18 + 10):} stage 1 in FP32 for 117M bytes, stage 2 ternary for 13M
        (90/10 of the 130M budget), at stage-2 LR $1\!\times\!10^{-4}$ and at the swept value; plus a
        five-point stage-2 LR sweep per architecture at seed 0. The transformer arm here is the
        $1\times52$ shape, not the pre-specified $3\times32$, and its from-scratch reference uses
        the same shape.
\end{itemize}
Parameter-matched widths come from a depth schedule fixed before any results were seen
(\texttt{configs/}\allowbreak\texttt{scales.json}); the shape sweep then tests that choice directly.

\begin{table}[h]\centering\small
\caption{Run accounting. Two of the tuned FP-init arms' seed-0 runs are the corresponding
sweep runs reused rather than retrained, so 100 run directories correspond to 98 distinct
trainings.}\label{tab:runs}
\begin{tabular}{lrrr}\toprule
Experiment & cells & seeds & runs \\\midrule
Main grid (routed, $3\times32$ transformer) & 8 & 3 & 24 \\
Shape sweep (5 shapes @16M, 3 @130M) & 12 & 3 & 36 \\
Gated SSM & 4 & 3 & 12 \\
FP-init arms (2 architectures $\times$ 2 stage-2 LRs, incl.\ stage 1) & 6 & 3 & 18 \\
Stage-2 LR sweep & 10 & 1 & 10 \\\midrule
\textbf{Total} & & & \textbf{98 distinct} \\\bottomrule
\end{tabular}
\end{table}

\begin{table}[h]\centering\small
\caption{Exact configurations. Every model ternarizes its token embedding, unembedding and all
weight matrices (plus the depthwise conv and router, for the routed block). ``FP kept'' is the
share of parameters that remain full precision: per-channel SSM scalars and LayerNorms everywhere,
plus the transformers' learned positional embeddings, which dominate that column.}\label{tab:spec}
\begin{tabular}{llrrl}\toprule
System & configuration & params & FP kept & what stays full precision \\\midrule
Routed 3-pathway & $d{=}64$, 4 layers, $d_{\text{head}}{=}16$, top-$k{=}4$ & 60{,}800 & 2.3\% & SSM scalars, LayerNorms \\
Gated SSM (ours) & $d{=}44$, 3 layers, $d_{\text{ff}}{=}2d$ & 64{,}196 & 1.6\% & SSM scalars, LayerNorms \\
Transformer $1\times52$ & $d{=}52$, 1 layer, $d_{\text{ff}}{=}2d$, 4 heads & 61{,}880 & 22.0\% & + positional embedding \\
Transformer $2\times36$ & $d{=}36$, 2 layers, $d_{\text{ff}}{=}4d$, 4 heads & 59{,}112 & 16.2\% & + positional embedding \\
Transformer $3\times32$ & $d{=}32$, 3 layers, $d_{\text{ff}}{=}4d$, 4 heads & 61{,}888 & 14.0\% & + positional embedding \\
Transformer $4\times28$ & $d{=}28$, 4 layers, $d_{\text{ff}}{=}4d$, 4 heads & 59{,}640 & 12.9\% & + positional embedding \\
Transformer $6\times24$ & $d{=}24$, 6 layers, $d_{\text{ff}}{=}4d$, 4 heads & 60{,}528 & 11.2\% & + positional embedding \\\bottomrule
\end{tabular}

\vspace{2pt}\footnotesize The transformers are within 3\% of the routed model's parameter count;
the gated SSM is 5.6\% larger, which flatters it and should be read as a caveat on §6.2. The
$1\times52$ shape uses $d_{\text{ff}}{=}2d$ because $4d$ at that width overshoots the budget.
\end{table}

\paragraph{Fixed recipe.} All runs use the published 944K recipe: AdamW
($\beta_1{=}0.9$, $\beta_2{=}0.95$, weight decay 0.1 on all parameters), peak LR
$3\!\times\!10^{-4}$ with cosine decay to $3\!\times\!10^{-5}$, gradient clipping 1.0,
sequence length 256, batch $64\times4$ (65{,}280 predicted tokens per step). Warmup is
3.33\% of steps (the published $1000/30000$), and each budget is a separate run whose schedule
ends at that budget.

\paragraph{Data.} TinyStories V2 \citep{eldan2023tinystories} train and validation files
concatenated as raw bytes: 2{,}250{,}255{,}763 bytes (SHA-256 recorded in a manifest in the
repository), cut into 256-byte chunks, 10\% held out, leaving 2.03B training tokens. Both budgets
are therefore well under one epoch (16M $=$ 0.008 epochs, 130M $=$ 0.064), so no result here is
affected by repeated data. The held-out chunks are drawn at random from the same corpus, so
near-duplicate stories could straddle the split; since every system is evaluated on the same
slice this cannot favour one architecture, but absolute losses may be slightly optimistic.

\paragraph{Metric and decision rule.} The primary metric is validation loss in nats/byte of
the \emph{last} checkpoint on 256 held-out batches (best-on-validation selection is
available in the released logs). For each comparison we report the mean over seeds, the relative gap, and a
noise floor $\sigma=\sqrt{\sigma_a^2+\sigma_b^2}/\mu_b$; a gap is called real only if it
exceeds $2\sigma$, and comparisons with fewer than two seeds per group are reported as
undetermined. Every ternary run is compared to a full-precision anchor trained on the same
total number of bytes. Initialization and data-order seeds are controlled separately.

\paragraph{Diagnostics.} For all routed runs we log per-layer router entropy, maximum
router probability and mean pathway weights; for all quantized runs we log the relative $L_1$
change of shadow weights from their reference, the fraction of ternary levels that changed,
and the zero fraction. These are used for the router analysis in \S\ref{sec:router} and are
otherwise available in the released logs; we do not report them run by run.

\paragraph{Availability.} Training code, configs, the corpus manifest and the per-run logs are
at \url{https://github.com/veldanda/ByteLM} under release tag \texttt{p1-v1}; every table and
figure in this paper is regenerated from those logs by \texttt{experiments/analyze\_all.py}.

\paragraph{Compute.} Every run was trained on one Apple MacBook Air (M2, 16\,GB, PyTorch 2.8,
MPS backend), ${\sim}40$ device-hours in total, at zero marginal cost.

\section{Results}\label{sec:results}
All numbers are validation loss in nats/byte on 256 held-out batches of the last checkpoint,
mean $\pm$ sd over three seeds. The models are byte-level, so a training ``token'' is one byte and
we state budgets in bytes: 16M and 130M bytes, i.e.\ ${\sim}260$ and ${\sim}2{,}140$ bytes per
parameter. Budgets are matched within every comparison. Note that changing the budget changes three
things at once: data seen, optimizer steps (245 vs.\ 1{,}991) and schedule length, so ``budget''
here means all three together, not data quantity alone.

\begin{table}[h]\centering\small
\caption{All systems at 60K parameters. Ternary penalty = (ternary $-$ FP)/FP.}\label{tab:main}
\begin{tabular}{lcccc c}\toprule
 & \multicolumn{2}{c}{16M bytes} & \multicolumn{2}{c}{130M bytes} & \\
System & FP32 & ternary & FP32 & ternary & penalty @130M \\\midrule
Gated SSM (ours) & \textbf{2.704} $\pm$ .041 & 3.376 $\pm$ .093 & \textbf{1.516} $\pm$ .012 & \textbf{1.942} $\pm$ .018 & +28.1\% \\
Routed 3-pathway (Atome) & 2.808 $\pm$ .062 & \textbf{3.290} $\pm$ .065 & 1.668 $\pm$ .024 & 1.993 $\pm$ .032 & +19.5\% \\
Transformer $1\times52$ & 2.835 $\pm$ .016 & 3.426 $\pm$ .034 & 2.195 $\pm$ .003 & 2.311 $\pm$ .003 & \textbf{+5.3\%} \\
Transformer $2\times36$ & 3.080 $\pm$ .031 & 3.752 $\pm$ .048 & 2.169 $\pm$ .016 & 2.320 $\pm$ .002 & +7.0\% \\
Transformer $3\times32$ & 3.239 $\pm$ .152 & 3.870 $\pm$ .094 & 2.143 $\pm$ .028 & 2.328 $\pm$ .004 & +8.6\% \\
Transformer $4\times28$ & 3.411 $\pm$ .070 & 4.103 $\pm$ .044 & -- & -- & \\
Transformer $6\times24$ & 3.477 $\pm$ .114 & 4.162 $\pm$ .052 & -- & -- & \\\bottomrule
\end{tabular}
\end{table}

\begin{figure}[h]\centering
\pgfplotsset{every axis title/.style={above,at={(0.5,1.0)},font=\small},
  every axis/.append style={width=0.46\textwidth,height=5.4cm,grid=major,
  xlabel={transformer depth (layers)},tick label style={font=\footnotesize},
  label style={font=\footnotesize},clip=false}}
\begin{tikzpicture}
\begin{axis}[ylabel={val loss (nats/byte)},title={16M bytes},xtick={1,2,3,4,6},
             ymin=2.60,ymax=3.62,xmin=0.6,xmax=6.9]
\addplot[color=blue,mark=*,error bars/.cd,y dir=both,y explicit] coordinates
  {(1,2.8350)+-(0,0.016) (2,3.0797)+-(0,0.031) (3,3.2386)+-(0,0.152) (4,3.4107)+-(0,0.070) (6,3.4767)+-(0,0.114)};
\addplot[dashed,thick,domain=0.6:6.9,forget plot]{2.8080};
\addplot[dotted,thick,domain=0.6:6.9,forget plot]{2.7041};
\node[font=\scriptsize,anchor=west] at (axis cs:3.6,2.86) {routed};
\node[font=\scriptsize,anchor=west] at (axis cs:3.6,2.665) {gated SSM};
\node[font=\scriptsize,anchor=west,blue] at (axis cs:4.2,3.53) {transformer};
\end{axis}
\end{tikzpicture}\hfill
\begin{tikzpicture}
\begin{axis}[title={130M bytes},xtick={1,2,3},ymin=1.44,ymax=2.30,xmin=0.75,xmax=3.4]
\addplot[color=blue,mark=*,error bars/.cd,y dir=both,y explicit] coordinates
  {(1,2.1950)+-(0,0.003) (2,2.1692)+-(0,0.016) (3,2.1434)+-(0,0.028)};
\addplot[dashed,thick,domain=0.75:3.4,forget plot]{1.6677};
\addplot[dotted,thick,domain=0.75:3.4,forget plot]{1.5164};
\node[font=\scriptsize,anchor=west] at (axis cs:1.9,1.715) {routed};
\node[font=\scriptsize,anchor=west] at (axis cs:1.9,1.562) {gated SSM};
\node[font=\scriptsize,anchor=west,blue] at (axis cs:1.6,2.23) {transformer};
\end{axis}
\end{tikzpicture}
\caption{Full-precision validation loss at 60K parameters, three seeds (bars: sd). Every
transformer point is within 3\% of the same parameter count; depth, width, FFN ratio and
positional-embedding share vary together along the $x$-axis (Table~\ref{tab:spec}).
\textbf{Left:} at a short budget the best shape (1 layer, width 52) ties the routed block while the
worst (6 layers, width 24) is 22.6\% worse than the best and 19.2\% worse than the routed block. \textbf{Right:} at 130M bytes the ordering inverts (depth
now helps) and both recurrent architectures pull clearly ahead, the gated SSM furthest.
Ternary runs (Table~\ref{tab:main}) show the same shape ordering.}\label{fig:shape}
\end{figure}
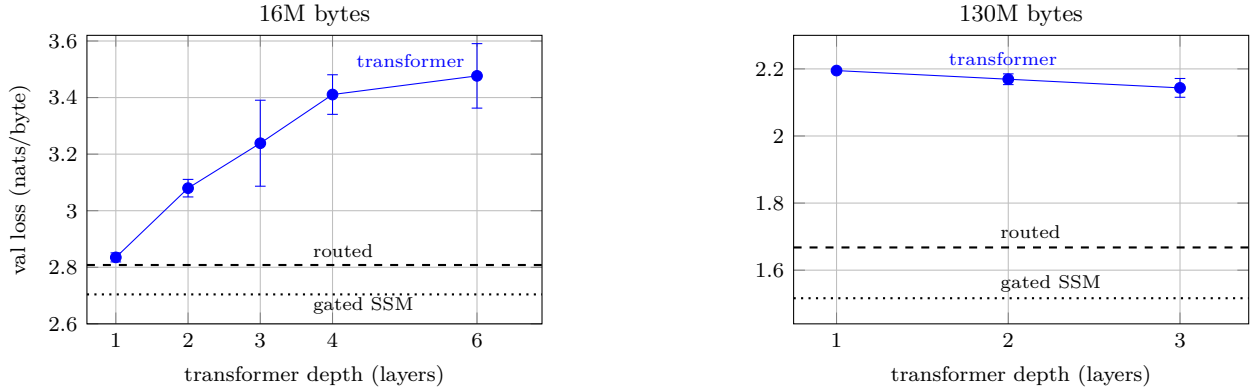

\subsection{Baseline shape decides the verdict}
At 16M bytes, param-matched transformer shapes span 2.835--3.477 nats/byte in full precision:
the worst shape is 22.6\% worse than the best (equivalently, the best is 18.5\% better), a span
far wider than any architecture effect we measure at this budget: the routed model is 0.95\% and
the gated SSM 4.6\% better than the best shape (Figure~\ref{fig:shape}). Against the best shape ($1\times52$) the routed model is $0.95\%$ better in full precision
($2\sigma_{\text{rel}}=4.5\%$) and $4.0\%$ better in ternary ($2\sigma_{\text{rel}}=4.3\%$),
both inside the noise band. Against the worst shape ($6\times24$) it is $19.2\%$ and $21.0\%$
better, far outside it. A study that
fixes one shape can therefore report either verdict.

One caveat on the best shape itself: $1\times52$ differs from the other shapes in more than depth
and width. Its feed-forward ratio is $2d$ rather than $4d$ ($4d$ overshoots the parameter budget at
that width), and its learned positional embedding occupies 22\% of its parameters against 11--16\%
for the deeper shapes (Table~\ref{tab:spec}). ``Shape'' here therefore means the whole
depth/width/FFN/position-budget package, not depth alone.

Two of these comparisons are exploratory rather than confirmatory: routed vs.\ the best observed
shape, in full precision and in ternary. The pre-specified baseline was
$3\times32$, from a depth schedule fixed before any runs; the other four shapes were added after
seeing that result, and ``routed vs.\ the best observed shape'' is therefore a post-hoc descriptive
comparison, reported as such.

Crucially the ordering \emph{reverses with budget}: at 130M bytes the shallow-wide $1\times52$
shape is the worst transformer (2.195) and the deeper $3\times32$ the best (2.143). Shape must be
swept per budget; it cannot be chosen once.

\subsection{At a larger budget the gain is real, but it does not require routing}
At 130M bytes the routed model beats each of the three transformer shapes evaluated at that budget
($1\times52$, $2\times36$, $3\times32$) by 22.2--24.0\% in full precision and 13.7--14.4\% in
ternary, all outside the noise band. The two
deepest shapes were not rerun at 130M, so this is a claim about the three strongest shapes at 16M,
not about the whole shape family.

Our gated diagonal-SSM block (a single recurrent pathway with a GELU feed-forward layer and no
router) beats the routed model by 9.1\% in full precision ($2\sigma_{\text{rel}}=3.2\%$) and
matches it in ternary ($-2.6\%$, $2\sigma_{\text{rel}}=3.7\%$, within the band). Routing is
therefore not required for the advantage: a much simpler gated recurrence reaches comparable or
better loss at the same parameter and byte budget. We do \emph{not} claim to isolate recurrence as
the causal ingredient (our block also differs from the routed one in having a feed-forward layer
and denser channel mixing), only that the router is not what buys the gain.

\subsection{The ternary penalty differs by architecture, with caveats we cannot yet remove}
At 130M bytes the ternary penalty is $+5.3\%$ for the best transformer shape, $+19.5\%$ for the
routed model and $+28.1\%$ for the gated SSM; in absolute terms $0.116$, $0.326$ and $0.425$
nats/byte. The transformer's penalty falls sharply with budget ($+20.9\% \to +5.3\%$), matching what
ParetoQ reports at $\geq$125M parameters, while the recurrent models' do not ($+24.9\% \to +28.1\%$
and $+17.2\% \to +19.5\%$). Because the 16M penalties sit inside noise bands of
$2\sigma_{\text{rel}}=7.5\%$ and $6.4\%$, the defensible statement is that the recurrent penalties
\emph{do not shrink}, not that they grow. The architecture gap consequently narrows without
closing: the routed model is $22.2\%$ better than the best transformer shape in full precision and
$13.7\%$ better in ternary.

Three alternative explanations are not excluded, and we flag them rather than claim the result is
clean. \textbf{Unequal quantization.} Our transformers keep learned positional embeddings in full
precision ($11$--$22\%$ of their parameters, largest for the shallow-wide shape) against
$1.6$--$2.3\%$ for the recurrent models (Table~\ref{tab:spec}). A model that quantizes less of
itself should lose less, so some unknown share of the difference is bookkeeping rather than
architecture. A replication with a zero-parameter position scheme (RoPE, ALiBi, sinusoidal) would
settle it. \textbf{An untuned ternary baseline.} Every from-scratch ternary run uses the
full-precision recipe's peak learning rate; \S\ref{sec:fpinit} shows the ternary stage of the
FP-init arm prefers a rate $10\times$ larger, so the from-scratch penalties reported here may be
inflated, possibly unequally across architectures. \textbf{Loss level rather than architecture.}
At 130M bytes the recurrent models sit at 1.52--1.67 nats/byte and the transformers at 2.14--2.20,
so penalty and loss level covary; at 16M, where the routed model (2.808) and the $1\times52$
transformer (2.835) reach similar loss, their penalties are also similar (17.2\% vs.\ 20.9\%). The
penalty may track how much fine structure a model has already fitted rather than what kind of model
it is. A loss-matched comparison would separate these, and we have not run one.

\subsection{A 90/10 FP-then-ternary schedule helps, at a learning rate 10$\times$ the pretraining peak}\label{sec:fpinit}
Following ParetoQ we ran 90\% of the budget in full precision and the last 10\% in ternary,
token-matched to the from-scratch arm (Table~\ref{tab:fpinit}). With a conventional fine-tuning
rate ($1\!\times\!10^{-4}$) the recipe looks clearly \emph{worse} than training ternary from
scratch. A five-point stage-2 sweep (seed 0) shows loss falling monotonically up to
$3\!\times\!10^{-3}$, an order of magnitude above the pretraining peak of $3\!\times\!10^{-4}$;
at that rate the same recipe \emph{wins}. The sweep had not saturated at its upper end, so $3\!\times\!10^{-3}$ is a lower bound on the
useful rate, not an estimated optimum; we did not extend it further, and where the recipe becomes
unstable is untested. Note also what this comparison is and is not: it tests a 90/10 allocation of
one byte budget between full-precision and ternary optimization, not quantization-aware
fine-tuning in general. That so short an adaptation phase (10\% of the budget) prefers so large a
step size suggests the ternary stage is doing something closer to re-solving the weights than to
fine-tuning them, which is consistent with the ``reconstruction'' behaviour ParetoQ reports at
larger scale.

\begin{table}[h]\centering\small
\caption{FP-init (90\% FP $\to$ 10\% ternary) vs.\ ternary from scratch, token-matched at 130M.
Left: stage-2 LR sweep, seed 0. Right: three-seed arms. The transformer arm is the $1\times52$
shape in both the FP-init runs and the from-scratch reference they are compared against, so the
comparison holds shape fixed.}\label{tab:fpinit}
\begin{tabular}{lccccc @{\hskip 1em} lccc}\toprule
stage-2 LR & 3e-5 & 1e-4 & 3e-4 & 1e-3 & 3e-3 & arm (3 seeds) & loss & vs.\ scratch & $2\sigma_{\text{rel}}$ \\\midrule
routed  & 2.758 & 2.335 & 2.110 & 1.944 & \textbf{1.911} & routed, 1e-4 & $2.299\pm.041$ & $+15.3\%$ & 5.2\% \\
transformer & 2.701 & 2.474 & 2.348 & 2.298 & \textbf{2.265} & routed, 3e-3 & $1.900\pm.010$ & $\mathbf{-4.7\%}$ & 3.4\% \\
 & & & & & & transf., 1e-4 & $2.475\pm.014$ & $+7.1\%$ & 1.2\% \\
 & & & & & & transf., 3e-3 & $2.262\pm.003$ & $\mathbf{-2.1\%}$ & 0.4\% \\\bottomrule
\end{tabular}
\end{table}

\subsection{What the router actually does}\label{sec:router}
The routing diagnostics logged for every routed run support the same conclusion from inside the
model. At the end of a 130M-byte ternary run, the mean per-token mixing weight on the state (SSM)
pathway is 0.49--0.73 across the four layers, against 0.09--0.25 for the local convolution and
0.19--0.28 for sparse attention; the full-precision run is more extreme still (state 0.54--0.79).
Per-layer router entropy stays at 0.67--0.87 nats against a maximum of $\ln 3 = 1.099$, so the
router neither collapses to one pathway nor stays uniform: it settles on a mixture dominated by
the recurrence. A model whose router spends most of its weight on the SSM pathway is consistent
with a plain gated SSM matching it.

\subsection{On the original H-budget hypothesis}
We had hypothesised that the published 60K advantage would shrink as the training budget grew
(the two published points differ by ${\sim}650\times$ in training bytes). It does not: against the shape
the original work used, the routed model's advantage \emph{grows} from $13.3\%$ at 16M bytes to
$22.2\%$ at 130M bytes. The confound we identified is real, but it is not what explains the result; at our two
controlled budgets, baseline shape is.

\subsection{What the experiments do and do not establish}
\begin{table}[h]\centering\small
\caption{Claims against the evidence that supports them.}\label{tab:claims}
\begin{tabular}{@{}p{6.5cm}p{4.0cm}p{4.6cm}@{}}\toprule
Claim & Evidence & Status \\\midrule
Baseline depth/width choice can decide an architecture verdict at 60K & 5-shape sweep, 2 budgets & supported \\
The published 60K advantage is explained by training budget (H-budget) & 16M vs.\ 130M bytes & not supported \\
Routing is required for the routed block's advantage & gated SSM comparison & not supported \\
A simple gated recurrence suffices to match or beat it & gated SSM, 2 budgets & supported \\
Recurrence \emph{per se} is the causal ingredient & not isolated (FFN, mixing differ) & untested \\
The ternary penalty differs by architecture at 130M & 3 families, 2 budgets & supported \\
The recurrent models' ternary penalty \emph{grows} with budget & 16M values inside noise bands & not established \\
\dots{} and is attributable to architecture alone & transformers keep 11--22\% of params in FP; ternary baseline untuned & confounded \\
A 90/10 FP$\to$ternary schedule beats\newline all-ternary training & 3 seeds, tuned stage-2 LR & supported, vs.\ an untuned\newline from-scratch baseline \\
$3\!\times\!10^{-3}$ is the optimal stage-2 LR & sweep stops there & not established \\
Any of this holds beyond 60K parameters or beyond TinyStories & -- & unknown \\\bottomrule
\end{tabular}
\end{table}

\section{Limitations}\label{sec:limits}
\textbf{One parameter count.} Every result is at 60K parameters. We cannot say whether the
architecture ordering or the ternary penalties change at 1M or 10M, and we therefore do not
address the 944K reversal that motivated the audit. Training a 944K grid to a comparable
bytes/parameter budget is roughly two orders of magnitude more compute than a laptop allows.

\textbf{One corpus, and a byte-level one.} TinyStories favours local, repetitive structure, and
byte-level input makes short-range dependencies dominant; both plausibly flatter convolutional and
recurrent pathways relative to attention. The architecture $\times$ precision interaction we report
is established on TinyStories V2 at byte level; whether it persists on less locally structured
corpora (code, multilingual text, subword tokenization) is open, and the specific gap percentages
should not be carried over. The tokenization itself is part of this: English text is roughly four
bytes per subword token, so a byte-level model sees sequences several times longer at the same
context width, which weights local sequential structure more heavily and plausibly favours the
convolutional and recurrent pathways over attention. A subword-tokenized replication could narrow
the architecture gaps we report even if their ordering holds. A wider-domain corpus is the single most valuable follow-up.

\textbf{Two budgets, low bytes/parameter at the top end.} 16M and 130M bytes
(${\sim}260$ and ${\sim}2{,}140$ bytes/parameter); the shortest budget is only 245 optimizer steps.

\textbf{Scope of the baselines.} The transformer shapes are param-matched within 3\% but share one
recipe; we did not tune learning rate per shape. The gated SSM is deliberately simple and is not a
Mamba-class model. Loss is the only quality metric. Wall-clock numbers come from a laptop, not an
MCU; deployment cost is left to follow-up work.

\textbf{Unequal quantization across architectures.} Our transformers keep their learned
positional embeddings in full precision ($11$--$22\%$ of their parameters) against
$1.6$--$2.3\%$ for the recurrent models, so the ``ternary'' arms are not equally quantized. Any
architecture-dependent ternary penalty we report is therefore an upper bound on the architectural
component; a replication with a zero-parameter position scheme is needed before the difference can
be attributed to architecture.

\textbf{The from-scratch ternary baseline is untuned.} Every from-scratch ternary run reuses the
full-precision peak learning rate. The FP-init sweep shows that ternary optimization prefers a rate
an order of magnitude larger, so the from-scratch ternary losses (and hence every ternary
penalty in this paper) may be inflated, possibly by different amounts across architectures. A
per-architecture ternary learning-rate sweep is the cheapest experiment that would tighten
\S\ref{sec:results}.

\textbf{Stage-2 LR not bracketed.} The FP-init sweep had not saturated at $3\!\times\!10^{-3}$,
so the reported FP-init results are a lower bound on what the recipe can do.

\section{Conclusion}
A reader might take three things from the published comparison: that the routed block beats a
parameter-matched transformer at 60K, that its router is what produces the advantage, and that the
cost of ternary weights is a property of precision rather than of architecture. At 60K parameters
the first two do not survive a controlled re-run. The advantage at a short budget is at least partly a baseline-shape artifact,
and the advantage that remains at a longer budget does not require routing: a single gated
recurrence does at least as well, though our comparison does not isolate recurrence itself as the
cause. What survives is more qualified: the cost of ternary weights differs sharply by architecture at the
larger budget (the transformer recovers with more training, the recurrent models do not), though
our transformers keep 11--22\% of their parameters in full precision and our from-scratch ternary
baseline was never learning-rate tuned, and loss level covaries with architecture here, so we report that difference as an observation
with three live alternative explanations rather than as an established architectural law. The cleanest new result is
the recipe one: a 90/10 full-precision-then-ternary schedule needs a learning rate far above the
usual fine-tuning range before it beats all-ternary training at all, though it is measured
against a from-scratch baseline that was never itself learning-rate tuned, so the size of the win,
and not only its sign, is provisional. None of this establishes behaviour beyond 60K
parameters or beyond TinyStories. Both of the methodological traps we hit (a fixed
baseline shape, an untuned stage-2 learning rate) individually reverse a headline result, which is
worth more caution than the field currently gives them at this scale.

What this does not settle is the scaling question that motivated the audit. We show the published
60K comparison does not support the interpretation placed on it; the reported reversal at 944K
parameters remains untested, because a comparable grid at that size is roughly two orders of
magnitude beyond a laptop budget. Whether the architecture ordering and the architecture-dependent
ternary penalty hold, shrink or invert between 60K and 1M parameters is the obvious next
experiment, and the one we would run first given compute.

\bibliographystyle{plainnat}
\bibliography{refs}

\appendix
\section{Exact faster kernels}\label{app:kernels}
\paragraph{Chunked-scan SSM.} The released implementation trains the diagonal SSM as a
depthwise convolution with a length-$L$ kernel ($O(L^2)$ per channel). We split the sequence
into chunks of $T{=}32$: within a chunk, the recurrence is a lower-triangular $T\times T$
matrix $M_{ts}=a^{t-s}$ applied per channel; across chunks the state is carried with $L/T$
sequential steps, $H_n = a^T H_{n-1} + h^{\text{end}}_n$, and added as $a^{t+1}H_{n-1}$.
Only non-negative powers $a^k$, $k\leq T$, appear, so the scheme is stable for any
$|a|<1$. Forward and all gradients agree with the original to $\leq2\times10^{-6}$ relative
error (also against the step-by-step recurrence), at 19$\times$ lower cost on CPU
(155\,ms $\to$ 8\,ms, $C{=}L{=}256$, batch 16). On the MPS backend the same substitution reduced
one block's SSM from 300\,ms to 14\,ms, measured while another training job shared the GPU, so
that ratio is indicative rather than a clean benchmark.
\paragraph{Gather top-$k$ attention.} Entries outside the top-$k$ are exactly zero after
softmax, so the output equals a softmax over the $k$ selected scores applied to the $k$
gathered value rows; this avoids scattering scores back into an $L\times L$ matrix.
Agreement $\leq1.2\times10^{-6}$. End-to-end, a toy training pipeline gives the same ternary--FP
relative gaps (+4.56\% vs.\ +4.57\%; +7.10\% vs.\ +7.11\%) under either implementation.
All runs in this paper use the fast path; \texttt{-{}-exact-kernels} restores the upstream
implementations for verification.

\end{document}